\documentclass[conference]{IEEEtran}
\IEEEoverridecommandlockouts    

\usepackage{multirow}
\usepackage{lipsum}
\usepackage{amsmath}
\usepackage{amssymb}
\usepackage[ruled,vlined]{algorithm2e}
\usepackage{graphicx}
\usepackage{booktabs}
\usepackage{arydshln}
\usepackage{xcolor}
\usepackage{hyperref}
\graphicspath{{./Figures/}}
\usepackage{tikz}
\usepackage{eso-pic}

\title{\LARGE \bf Generation of Vectorized Maps Beyond Vehicle View}

\author{
	\parbox{\textwidth}{%
		\centering
		Clara Gomez$^{1}$, Alberto Jaenal$^{2}$, Antonio Artuñedo$^{1}$, Jorge Godoy$^{1}$, Jorge Villagra$^{1}$%
	}%
	\thanks{$^{1}$  Centre for Automation and Robotics, CSIC-UPM, Spain.
		{\tt\small \{clara.gomez, antonio.artunedo, jorge.godoy, jorge.villagra\}@csic.es}
    }%
    \thanks{$^{2}$ Aragon Institute for Engineering Research (I3A), University of Zaragoza, Spain.
        {\tt\small ajaenal@unizar.es}
    }%
}

\begin{document}

\newcommand{\CG}[1]{\textcolor{black}{#1}}
\newcommand{\pending}[1]{\textcolor{blue}{#1}}
\newcommand{\name}{BeyondFormer}
	
	\maketitle
	\thispagestyle{empty}
	\pagestyle{empty}
	
\begin{abstract}


Autonomous driving relies on High Definition (HD) maps for safe navigation. Traditional HD maps construction is costly in hardware, data and human resources, which together with its update limitations hinders scalability. Recent works have proposed online alternatives for HD vectorized mapping from onboard sensors. However, sensor field of view is limited, and the range of the reconstructed maps ahead of the vehicle is insufficient for safe planning. 
\CG{This paper aims to address this limitation by proposing the novel \textit{beyond-view} vectorized map generation problem: given vectorized maps of the area sensed by the vehicle (\textit{in-view}), to generate plausible map continuations. To experimentally assess its feasibility, we propose \name, which, to the best of out knowledge, is the first work designed towards beyond-view map generation. Given the novelty of the problem, we generate the first dataset specifically designed for it and evaluate the proposed approach. The results demonstrate consistent performance across diverse scenarios, establishing learning-based methods as a promising direction for map forecasting in autonomous driving. Beyond demonstrating the feasibility of the task, we provide an extensive discussion of the method's limitations and identify key future research directions for scaling it to more complex driving conditions.} Code is available at \url{https://git-autopia.car.upm-csic.es/beyondformer}.


\end{abstract}

\AddToShipoutPictureFG*{%
  \begin{tikzpicture}[remember picture,overlay]
    \node[
      draw=black,
      line width=0.4pt,
      inner xsep=6pt,
      inner ysep=4pt,
      text width=0.88\textwidth,
      align=center,
      font=\footnotesize
    ] at ([yshift=12mm]current page.south)
    {%
      \textcopyright\ 2026 IEEE. Personal use of this material is permitted.
      Permission from IEEE must be obtained for all other uses, in any current
      or future media, including reprinting/republishing this material for
      advertising or promotional purposes, creating new collective works,
      for resale or redistribution to servers or lists, or reuse of any
      copyrighted component of this work in other works.

      This paper has been accepted for publication at the
IEEE International Conference on Vehicular Electronics and Safety (ICVES) 2026.
    };
  \end{tikzpicture}%
}

\section{Introduction}
\label{sec:introduction}


Recent developments in autonomous driving have been boosted by High Definition (HD) maps \cite{seif2016autonomous}. HD maps are highly accurate maps that provide detailed and vectorized representations of road elements such as lanes and pedestrian crossings. 
They have become a widely adopted mapping solution enabling safe and reliable operation for commercial autonomous vehicles. 
However, traditional HD map construction is prohibitively expensive, requiring highly sensorized vehicles, time-consuming human annotations and costly data storage. They also require frequent updates as they quickly become outdated. These requirements make HD maps difficult to scale to large environments and long time horizons \cite{elghazaly2023high}.

Learning-based methods have recently emerged to 
reduce dependence on offline map creation 
by constructing online vectorized HD maps around the vehicle using onboard sensors \cite{li2022hdmapnet, luo2024augmenting}. These approaches 
are inherently limited by the sensor range and do not provide information beyond the observable region. In contrast, classical HD maps do not have this restriction \CG{enabling long-range planning}, which motivates studying whether road structure beyond the sensor field of view can be inferred from partial observations or \textit{in-view} maps. This is an unexplored gap for vectorized road representations.


\begin{figure}
    \centering
    \includegraphics[width=\columnwidth]{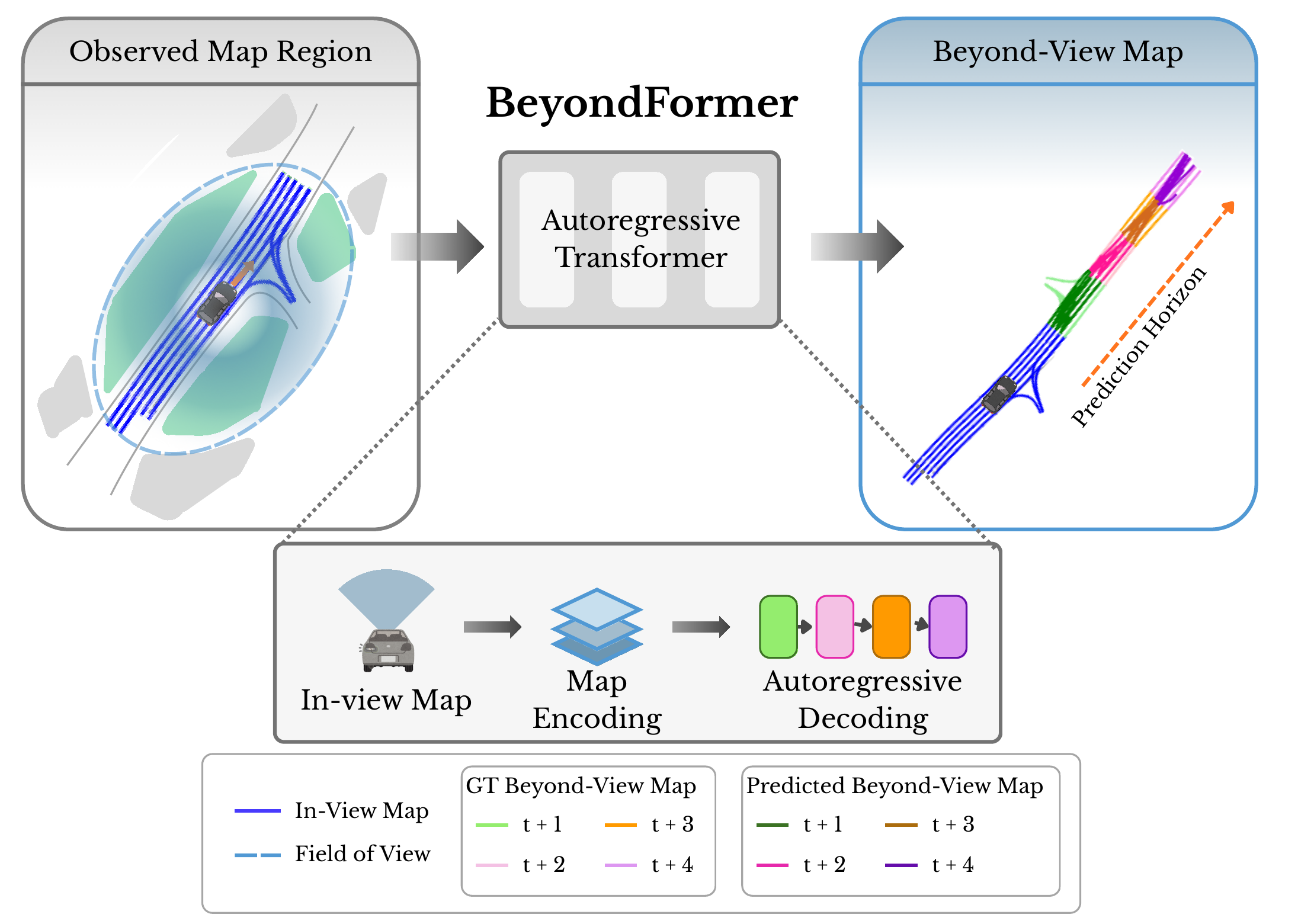}
    \vspace{-20pt}
    \caption{Given vectorized in-view maps of the observed region, \name~generates plausible continuations autoregressively.}
    \vspace{-15pt}
    \label{fig:intro}
\end{figure}


\CG{To address this gap, we formulate the \textit{beyond-view} vectorized map generation problem: consuming in-view maps to predict vectorized map continuations or beyond-view maps, as depicted in Fig.~\ref{fig:intro}. To experimentally assess this problem, we propose a learning based model, \name, which generates  plausible vectorized maps beyond the sensed area by the vehicle. This model adopts}
a \CG{multi-head} encoder-decoder architecture: a Transformer and a simple MLP are used to encode the in-view map and learn feature representations that are decoded with a Transformer decoder \CG{using a geometry head and an auxiliary topology head.} 



\name~is evaluated on map samples extracted from 3DHD CityScenes Dataset \cite{plachetka20223dhd}. Given the exploratory nature of this work, our goal is to introduce a novel problem formulation and establish a clear and reproducible \CG{first step} 
for beyond-view map generation,
rather than a highly optimized model. Accordingly, the experimental evaluation focuses on 
controlled settings and ablation studies, aiming to analyze the model behavior and identify key challenges and limitations.


To summarize, our contributions are:

\begin{itemize}
        \item The introduction and 
    study of the novel problem of 
    beyond-view vectorized map generation in autonomous driving.
    \item To the best of our knowledge, this is the first work to generate vectorized representations 
    of road environments beyond the sensed area by the vehicle, \CG{establishing a first step towards beyond-view map generation.} 
    \item Given the novelty of this work, no dataset matches the requirements for this problem. We 
    release a dataset with in-view and beyond-view map samples from 3DHD CityScenes Dataset
    , and evaluate \name~to analyze its capability for beyond-view map continuation.
\end{itemize}

\section{Related work}
\label{sec:relatedwork}

Current learning-based approaches for autonomous driving aim to reconstruct and forecast different aspects of the driving environment directly from sensor observations and structural priors reducing the dependence on offline maps. These works can be grouped into in-view vectorized map reconstruction and beyond-view trajectory and world model generation.

\textbf{In-view HD map reconstruction.} Most of the work on  HD map generation has been developed for in-view map prediction in which learning-based solutions transform on-board sensor readings (mainly images and point clouds) into vectorized Bird's-Eye View (BEV) representations \cite{li2022hdmapnet, qiao2023end, liu2023vectormapnet, liao2025maptrv2}. By encoding the scene as vectorized BEVs, the representation remains compact, memory-efficient, sensor-agnostic and invariant to viewpoint, supporting consistent environment modeling. HDMapNet \cite{li2022hdmapnet} pioneers in solving this problem, although they first obtain a rasterized map (which is highly resource demanding) to then extract the vectorized representation. VectorMapNet \cite{liu2023vectormapnet} is the first work to directly produce vectorized in-view maps. Of most interest for this work are previous works that, in addition to on board sensor readings, integrate SD maps as input for the reconstruction \cite{luo2024augmenting, jiang2024p, monninger2025navmapfusion, sun2025mind}. SMERF \cite{luo2024augmenting} encodes the SD map as a structure prior with a Transformer encoder that gets cross-attended with the observed evidence (BEV constructor) to reconstruct the in-view map. P-MapNet \cite{jiang2024p} and StreamMapNet \cite{yuan2024streammapnet} already identify that previous work prediction ahead of the vehicle is insufficient in distance, and point to more intelligent decisions when far-seeing solutions are available. However, they 
remain constrained to the observable region, extending the effective range of in-view reconstruction rather than predicting road structure beyond the sensor field of view. 
P-MapNet improves the long-range prediction by specific SD map and HD map encoders to aid the decoding process, and StreamMapNet by integrating a memory buffer that better captures the temporal dimension of the sequences to map. An interesting approach is also proposed in \cite{li2025amap} where in-view map prediction is produced with a teacher-student model in which the teacher model has privilege access to future information. Although the student model does not predict into the future, it benefits from ahead-looking capabilities.

\textbf{Beyond-view generation.} Generation of future views of the vehicle has been addressed to predict other users' movement through trajectory prediction \cite{zhang2024edge} and video or world model generation \cite{rowe2025scenario}. Works on the latter aim to forecast pixel-level scene evolution \cite{zheng2024occworld, yang2025geniedrive, yang2025driving, zhang2025epona}. However, these works do not explicitly model vectorized road geometry and topology beyond the observed region and,
as discussed by VectorMapNet, predicting pixels adds an extra computationally intensive process to the already demanding learning process of vectorization unsuitable for vectorized map generation.







Recently, similar ideas have been explored in indoor environments for structural scene completion and layout extrapolation \cite{ericson2024beyond, shah2025foresightnav}. However, transferring this idea to outdoor driving scenes introduces additional challenges due to significantly higher structural variability and topological complexity 
(multiple lanes and directions, roundabouts, lane splits, etc). Given the wide variability in driving situations, predicting beyond-view maps presents a significant challenge.

Although inspiration and technical solutions can be borrowed from the related work, to the best of the author's knowledge, this work is the first to study beyond-view vectorized map generation for autonomous driving.

\section{Method}
\label{sec:method}

\subsection{Problem formulation}

The problem of beyond-view vectorized map generation aims to infer the continuation of the road structure outside of the sensor field of view from a partial map observation. At each time step $t$, we assume that a perfect in-view map of the area perceived by the vehicle is built from sensor readings. These maps are represented as a set of vectorized instances, each instance containing a segment of a lane centerline. The complete set of lanes for the in-view map covers approximately $70$m around the vehicle (which aligns to the area that most in-view map generation methods provide \cite{monninger2025navmapfusion, sun2025mind}). The objective is \CG{to increase the vehicle's horizon by predicting}
plausible continuations of the centerlines in the driving direction, extending the map for another $100-150$m. 

\subsection{In-view input map and ground truth map}

\begin{figure}[b!]
    \centering
    \vspace{-15pt}
    \includegraphics[width=\columnwidth]{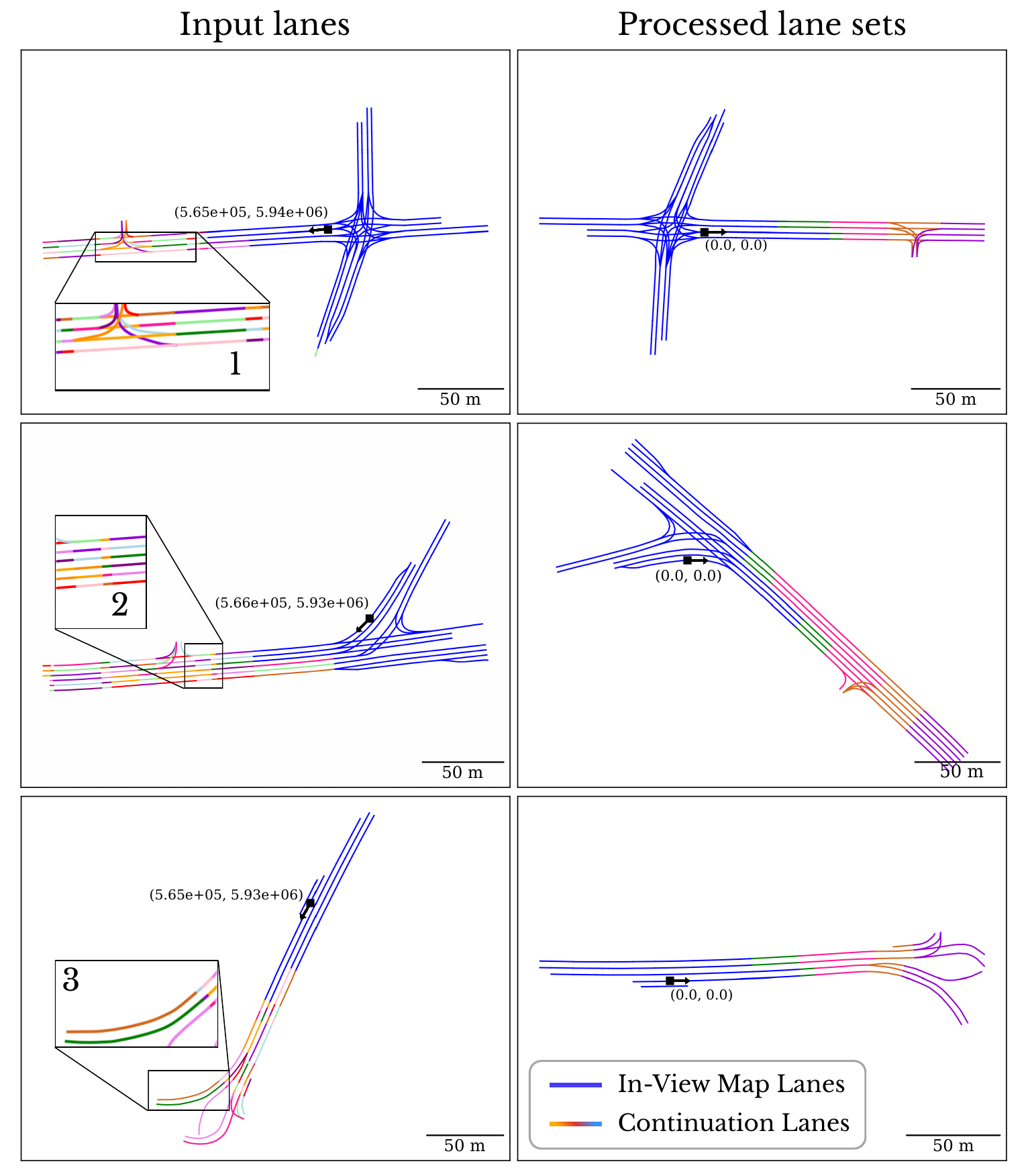}
    \vspace{-20pt}
    \caption{Examples of input data samples (left) and processed for autoregressive map prediction (right). \CG{In addition to the effect of normalization to vehicle pose, (1) shows unrealistically long lane segments, (2) shows multiple adjacent short lane segments and (3) shows imbalanced adjacent lane lengths. Processed lanes offer a more balanced and realistic representation.}}
    \vspace{-10pt}
    \label{fig:data_samples}
\end{figure}

\begin{figure*}[h]
    \centering
    \includegraphics[width=0.9\linewidth]{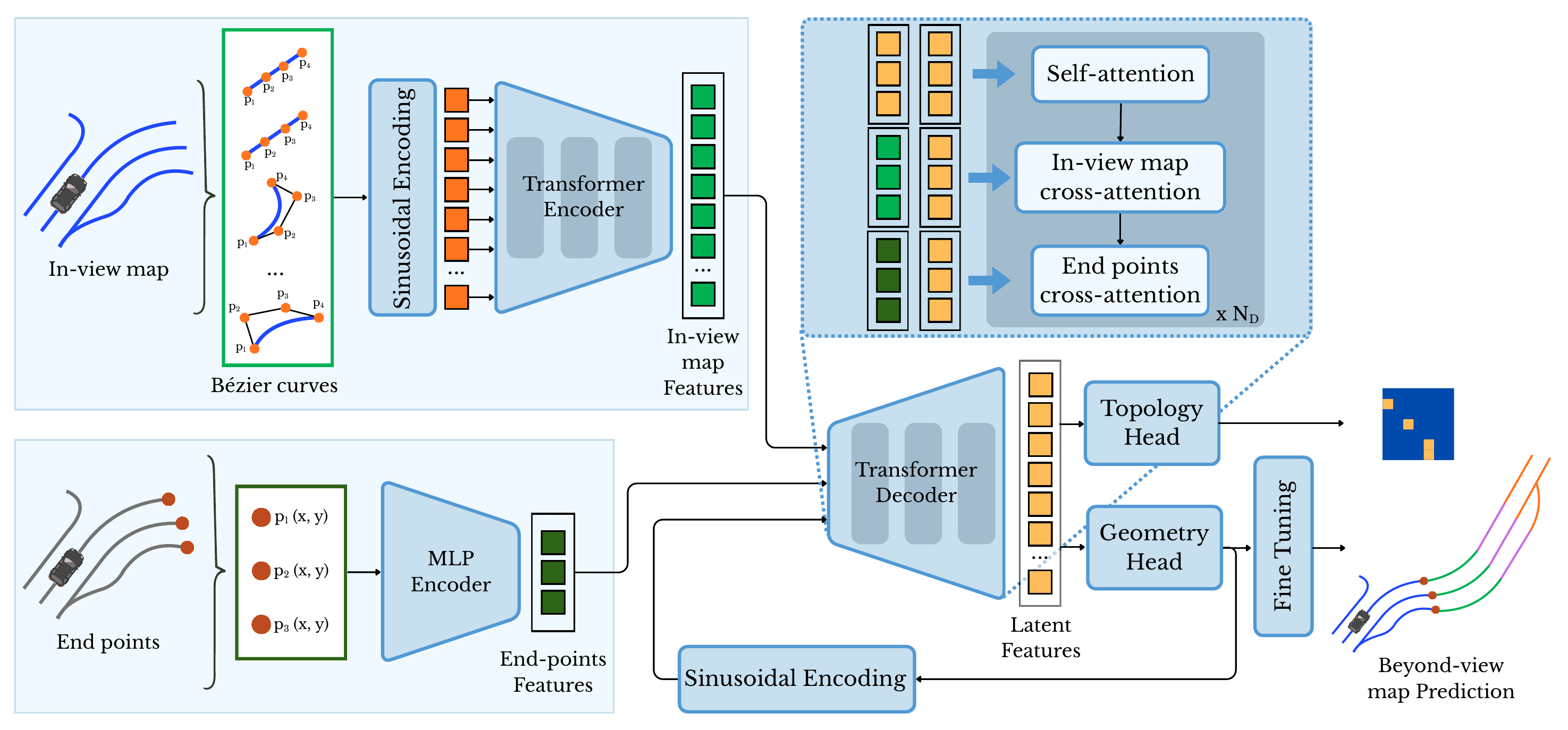}
    \vspace{-6pt}
    \caption{Architecture of \name comprising three main components: the in-view map transformer encoder that processes maps as B\'ezier curves, the end points encoder that uses a simple MLP encoder to extract end point features, and the \CG{decoding block formed by a transformer decoder with self-attention and cross-attention to the in-view map and end point features, a topology head and a geometry head that autoregressively generates the map continuation.}}
    \vspace{-10pt}
    \label{fig:method}
\end{figure*}

\CG{Every data sample in the in-view input map consists of the lane centerline segments in a $R$ radius around a global UTM coordinate (shown in blue in Fig.~\ref{fig:data_samples}). The ground-truth map consists of the lane centerline segments in a $R$ radius for the subsequent global coordinates of the trajectory up to $D$ distance. Extracted data samples are shown in the left column of Fig.~\ref{fig:data_samples}. We unify lane centerline segments lengths and normalize centerline points using the vehicle's position and orientation, resulting in the processed lanes shown in the right column of Fig.~\ref{fig:data_samples}.} \CG{Processed lanes solve several data problems as shown in the zoomed areas.}

Typically, lane centerline segments consist of a variable number of points depending on the complexity of the curve. As this is not suitable for learning-based algorithms, we turn each lane centerline segment into a cubic B\'ezier curve \cite{qiao2023end}. Cubic B\'ezier curves are convenient as they can represent curves up to two inflection points using only 4 control points.

The resulting in-view input map encompasses $M$ \CG{connected} B\'ezier curves, where each curve corresponds to a lane centerline segment. The resulting ground truth map encompasses $N$ sets of B\'ezier curves, where each set corresponds to the new centerline segments that the vehicle will observe in the next generation step, until the prediction horizon, $t+N$. The map is autoregressively generated using these sets.

\subsection{System Overview}

The pipeline of \name~is shown in Fig.~\ref{fig:method} and it consists of four building blocks: the in-view map encoder, the end points encoder, the \CG{multi-head} map decoder and fine tuning. 
\CG{ The input of our model are the in-view input map and the end points in the direction of travel from the in-view map. First we encode them into a feature representation, using a Transformer for the former and a MLP for the latter. Then, we apply cross-attention between both feature representations to decode a latent space through a transformer decoder. }
\CG{The latent space is processed with a geometry head to decode the beyond-view map and an auxiliary topology head to learn lane connectivity.} Up to here, the pipeline is trained end-to-end, afterwards an additional fine tuning step performs smoothing and connectivity checks to the predictions.

\subsection{In-view map Encoder}

A Transformer encoder processes the in-view map following the standard pipeline consisting of an input embedding, a positional encoding and $N_e$ encoder blocks with multi-head self-attention. The input embedding maps each of the four control points
$
\mathbf{P}_i^{\mathrm{in}}
=
\left\{\mathbf{p}_{i,j}^{\mathrm{in}}\right\}_{j=1}^{4}$ with $\mathbf{p}_{i,j}^{\mathrm{in}}
=
(x_{i,j},\,y_{i,j}).
$
of the $M$ B\'ezier curves into an embedding of size $d_{model}$. Then, we employ a sinusoidal positional encoding to map the sequentiality of the data. This enhances sensitivity to positional variations, enabling the model to effectively reason about the structure of the curves. The sinusoidal encoding is formulated based on the definition in \cite{vaswani2017attention}, as:

\begin{equation}
\begin{aligned}
&\mathrm{PE}(p, 2k) = \sin\!\left( p \cdot 10000^{- \frac{2k}{d_{\text{model}}}} \right)\\
&\mathrm{PE}(p, 2k+1) = \cos\!\left( p \cdot 10000^{- \frac{2k}{d_{\text{model}}}} \right)
\end{aligned},
\end{equation}

where $k$ is the point index in the embedding dimension and $p$ is the position in the input sequence.

The embedded data is then processed by the encoder blocks using $H$ self-attention heads to extract and encode the global geometry of the lanes in the in-view map.

\subsection{End points Encoder}

End points of the last lane segments in the direction of travel are used to guide the decoder to the coordinates to start the prediction. As for the in-view map, points 
$\mathbf{p}_{i}^{\mathrm{end}}
=
(x_{i},\,y_{i})
$
are transformed into embeddings of size $d_{model}$. Given the simplicity and low dimensionality of the end points, a 3-layer MLP encoder is suitable to build their feature representation.

\subsection{Multi-head Map Decoder}

The Transformer decoder produces \CG{a latent space of the lane map around the vehicle} from the encoded feature representations of the in-view map and the end points. The geometry head predicts beyond-view map sections and the auxiliary topology head predicts the connectivity between lanes. 

The decoder receives the encoder outputs (feature representations for in-view map and end points) and the input data for each generation step in $N$. The input data is processed by an input embedding and a sinusoidal positional encoding. $N_d$ decoder blocks with multi-head attention are used to produce the \CG{latent space}. First, self-attention is used to iteratively refine the prediction. Then, \name~fuses the in-view map features and the end point features with its own prediction by leveraging cross-attention. \CG{A 3-layer MLP is used as geometry head to produce the decoder output. The auxiliary topology head consists of a 3-layer MLP lane embedding and a 2-layer MLP pairwise edge logits. Edges are computed for every pair of lane features. The output of the auxiliary topology head is the learned adjacency matrix.} 
Finally, a projection layer transforms the decoder output from $d_{model}$ to B\'ezier curve points, 
$
\mathbf{P}_i^{\mathrm{out}}
=
\left\{\mathbf{p}_{i,j}^{\mathrm{out}}\right\}_{j=1}^{4}$ with $\mathbf{p}_{i,j}^{\mathrm{out}}
=
(x_{i,j},\,y_{i,j}).
$

The prediction is generated autoregressively, i.e., the model sequentially predicts the next lanes set given its own past predictions. We perform $N$ autoregressive generation steps according to the sets of data that we prepared from the ground truth. \CG{For example, in Fig.~\ref{fig:method} there are 3 of such steps.} A scheduled training \cite{bengio2015scheduled} strategy is used which gently changes the training process from a fully guided scheme using the true previous set, towards an unguided scheme which uses the generated set instead. This strategy has shown to be more robust to correct its own mistakes at inference. Practically, we start decoding using as input data a learned beginning-of-sequence element and then append the next ground-truth set or predicted set according to a scheduled-training ratio, $r_{st}$.

\subsection{Fine tuning}

Fine tuning is performed on the generated samples to ensure lane connectivity. Nearest neighbour matching is applied between generated lane starts and closest generated lane ends.

\subsection{Loss Function}

\CG{The loss function combines the geometry loss function, the topology loss function and a coupling loss:}

\vspace{-10pt}
\begin{equation}
\begin{aligned}
    \mathcal{L} &= \lambda_{geom}\mathcal{L}_{geom} + \lambda_{topo}\mathcal{L}_{topo} + \lambda_{coupling}\mathcal{L}_{coupling}~.
\end{aligned}
\end{equation}

\CG{The geometry loss function combines four losses:}


\vspace{-10pt}
\begin{equation}
\begin{aligned}
    \mathcal{L}_{geom} &= \lambda_{vec}\mathcal{L}_{vec} + \lambda_{dist}\mathcal{L}_{dist} + \lambda_{curv}\mathcal{L}_{curv}  + \lambda_{cont}\mathcal{L}_{cont}~,
\end{aligned}
\end{equation}


\noindent for which we discretize the ground truth and predicted B\'ezier curves in $K$ points. 
The vectorized Huber loss $\mathcal{L}_{vec}$ minimizes the error between matched centerline points. The distance loss $\mathcal{L}_{dist}$ supervises the accumulated length of each lane and its corresponding ground-truth lane. The curvature loss  $\mathcal{L}_{curv}$ supervises the difference in logarithmic curvature between the predicted lanes and ground truth. Curvature is calculated for the sampled points and accumulated over the lane. Logarithmic curvature is used to avoid loss explosion for straight lanes. 
The continuity loss $\mathcal{L}_{cont}$ supervises the connectivity between lane sections by minimizing the distance between each lane start point and the end points of any previous lane section. 

\CG{For the topology loss function $\mathcal{L}_{topo}$, we extract the ground truth adjacency matrix from the connected lanes and compute the binary cross-entropy with logits. The coupling loss $\mathcal{L}_{coupling}$ enforces consistency between the auxiliary head and the main head by minimizing the geometric end-to-start point distance between topological adjacent lane pairs.}

\section{Experimental evaluation}
\label{sec:designandimplementation}

This section validates \name, a proof of concept for beyond-view map generation, demonstrating its feasibility through quantitative and qualitative results.

\subsection{Experimental setup}

\begin{figure}
    \centering
    \includegraphics[width=\columnwidth]{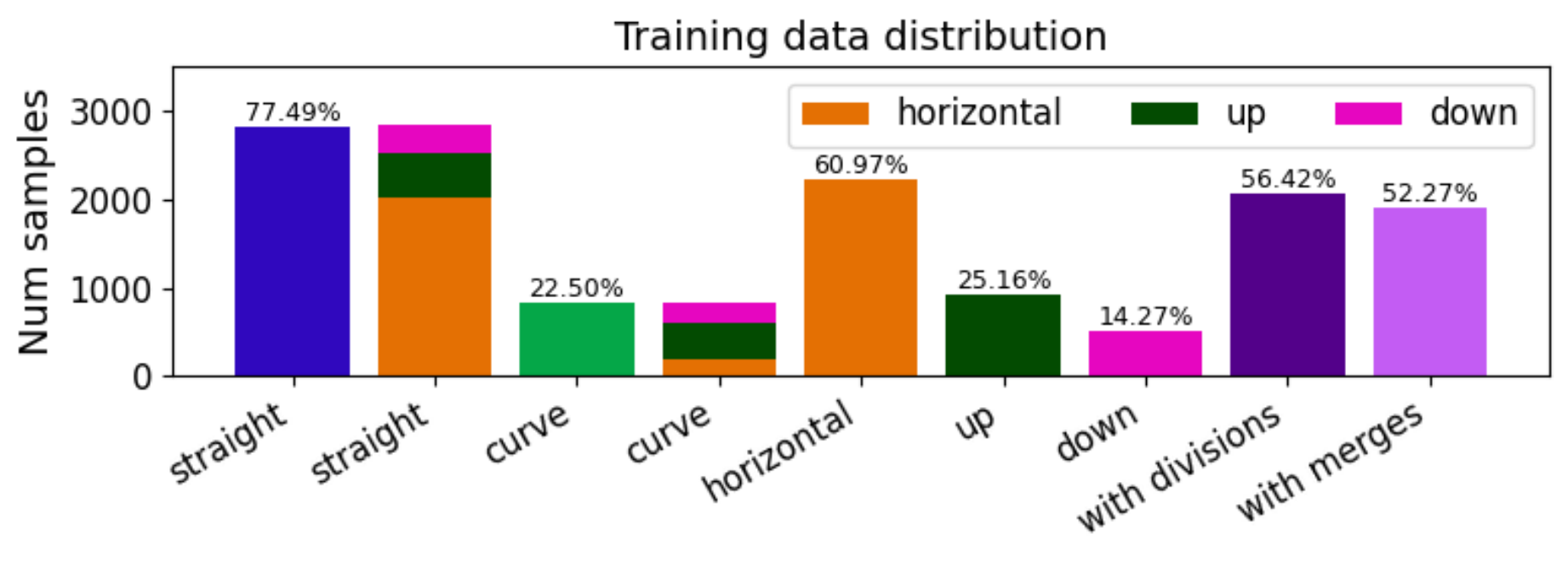}
    \vspace{-22pt}
    \caption{Distribution of road samples according to their direction (after normalization), shape and complexity.}
    \vspace{-13pt}
    \label{fig:data_distribution}
\end{figure}

\textbf{Dataset.} We evaluate \name~in 3DHD CityScenes Dataset \cite{plachetka20223dhd}, a real-world LiDAR-based 3D dataset of driving sequences in Hamburg. In addition to point clouds, it provides geo-localized vectorized representations of road elements. 3DHD CityScenes Dataset contains 57510 geo-localized samples for training, 8087 for validation and 13061 for test. The samples in each partition belong to disjoint road sections. We selected 3608 training, 171 validation and 104 test samples in simplified conditions (they do not contain roundabouts or intersections) for our experiments. We use $R=70$m as radius for the in-view and ground-truth map of each sample and $D=150$m as ground-truth distance. The distribution of road samples is presented in Fig.~\ref{fig:data_distribution}. Our processed dataset is available at \url{https://zenodo.org/records/22256829}.



\textbf{Hardware and training details.} Our model trains on 1 NVIDIA H200 GPU with a batch size of 64 using AdamW optimizer and an adaptative learning rate consisting in $2.5\%$ warm-up epochs and a cosine scheduler. Initial learning rate is set to $1 \times 10^{-4}$ and dropout to $0.1$. As samples consist of different lane numbers, we use zero padding and masking, and filtering of uninitialized lane predictions before metric computation. We train the model for 10k epochs with a generation horizon of $N = 4$ steps, and $H = 8$, $N_e = N_d = 6$, $d_{model} = 128$ and $r_{st} = 2 \times 10^{-4}$. \CG{The dimensions of MLPs are $[512, 256, 128]$ for the end points encoder, $[128, 128, 128]$ for the geometry head and for the topology head lane embedding and $[256, 128]$ for the edges.} The hyperparameters for loss calculation are: $K = 100$, \CG{$\lambda_{geom} = 1$, $\lambda_{topo} = 6$, $\lambda_{coupling} = 5$, $\lambda_{vec} = 1$, $\lambda_{dist} = 5$, $\lambda_{curv} = 350$ and $\lambda_{cont} = 50$. }\name~comprises $\sim 8M$ learnable parameters.


\CG{We provide a classical baseline by matching the endpoint tangent and second derivative (local curvature) of the previous segment while extending the B\'ezier curve by a fixed length.}

\textbf{Metrics.} We adopt common metrics from in-view map generation works \cite{luo2024augmenting, liu2023vectormapnet, jiang2024p}. We compute Root Mean Square Error (RMSE), Chamfer Distance (CD) and Fr\'echet Distance (FD) for discretized curves from the predicted and ground-truth B\'eziers. RMSE measures average pointwise deviation, CD measures average nearest-neighbor distance between two point sets, quantifying bidirectional geometric similarity, and FD measures the minimum coupling cost between two curves respecting the ordering of their points. 

\subsection{Beyond-view map generation performance}

\begin{figure}
    \centering
    \includegraphics[width=\columnwidth]{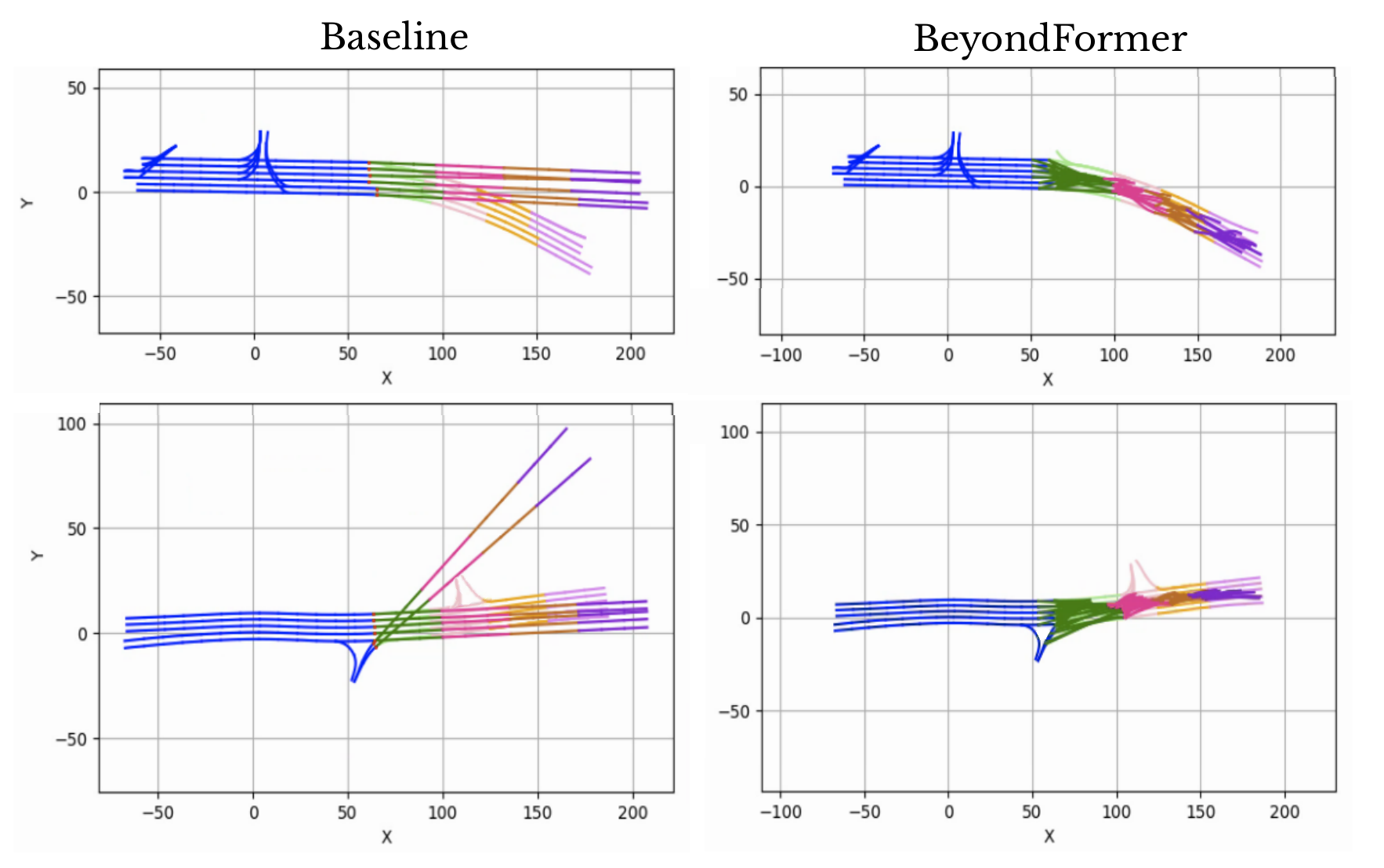}
    \vspace{-20pt}
    \caption{Examples of beyond-view map generation with \CG{baseline and} \name. The maps are color-coded as follows: blue for in-view maps, light colors (green, pink, orange, and purple) for ground-truth beyond-view maps, and dark colors for predicted outputs.}
    \label{fig:results}
    \vspace{-10pt}
\end{figure}

Qualitative results of beyond-view map generation using \CG{the baseline and} \name~are shown in Fig.~\ref{fig:results}. 
\CG{Cleaner lines are obtained with the baseline, although it is unable to predict curvature and direction changes. \name~achieves correct overall lane prediction in terms of curvature, length and distance, although failing to generate smooth parallel lanes. Both methods miss predictions of road splits.}

Tab.~\ref{tab:quantitative} and Tab.~\ref{tab:steps} show the quantitative performance of the method \CG{and the classical baseline} in the test data split. Tab.~\ref{tab:quantitative} provides mean $\pm$ standard deviation values for RMSE, CD and FD in meters. \CG{{\name~outperforms the baseline in terms of geometric precision with an overall average decrease in error of $3.14$m}. }
As predictions are generated for variable lane length and number, the percentage of error over the total prediction length (approx., $120$m and $4$ lanes) is also provided. Below $2\%$ error demonstrates the feasibility of the method, although absolute errors indicate room for improvement before operational performance. Additionally, Tab.~\ref{tab:steps} shows RMSE values for the 4 autoregressive steps. 
\CG{In both methods, errors increase with prediction horizon as each step relies on previous predictions, with a larger increase for the baseline}.

\begin{table}[t]
\centering
\caption{\name~quantitative evaluation}
\label{tab:quantitative}
\setlength{\tabcolsep}{2.5pt} 
\begin{tabular}{lccc}
\toprule
 & RMSE (m) $\downarrow$ & CD (m) $\downarrow$ & FD (m) $\downarrow$ \\
\midrule
Baseline & $8.78 \pm 2.16$ & $12.23 \pm 4.97$ & $10.80 \pm 2.79$\\
\noalign{\vskip 3pt}
\hdashline
\noalign{\vskip 3pt}
\name & $\mathbf{7.83 \pm 3.66}$ & $\mathbf{7.08 \pm 4.94}$ & $\mathbf{7.48 \pm 2.95}$\\
\name~($\%$)  & $1.4277$ & $1.2910$ & $1.3639$\\
\bottomrule
\end{tabular}
\end{table}

\begin{table}[t]
\centering
\caption{RMSE over Autoregressive steps (m) $\downarrow$}
\label{tab:steps}
\setlength{\tabcolsep}{2.5pt} 
\begin{tabular}{lcccc}
\toprule
& $t+1$ & $t+2$  & $t+3$ & $t+4$\\
\midrule
Baseline  & $\mathbf{3.22 \pm 0.64}$ & $8.12 \pm 1.52$ & $10.17 \pm 7.53$ & $12.72 \pm 3.75$\\
\name & $5.90 \pm 1.98$ & $\mathbf{6.34 \pm 2.34}$ & $\mathbf{7.34 \pm 4.28}$ & $\mathbf{9.06 \pm 6.89}$\\
\bottomrule
\vspace{-20pt}
\end{tabular}
\end{table}

\subsection{Ablation study}


We provide ablation studies for two key aspects in learning-based approaches: the loss function and model components. For the loss function ablation study, we consider only the geometry loss. The results in Tab.~\ref{tab:loss_ablation} show that RMSE, CD and FD can be reduced $23.60\%$, $30.66\%$ and $24.29\%$ respectively, by adding $\mathcal{L}_{dist}$, $\mathcal{L}_{curv}$ and $\mathcal{L}_{cont}$ to $\mathcal{L}_{vec}$. 

\begin{table}[b]
\centering
\caption{Ablation on Geometry Loss Function}
\label{tab:loss_ablation}
\begin{tabular}{lccc}
\toprule
Method & RMSE (m) $\downarrow$ & CD (m) $\downarrow$ & FD (m) $\downarrow$\\
\midrule
$\mathcal{L}_{vec}$ & $13.64 \pm 7.67$ & $10.47 \pm 7.12$ & $8.11 \pm 4.36$ \\
+ $\mathcal{L}_{dist}$  & $11.03 \pm 6.69$ & $8.18 \pm 6.74$ & $6.69 \pm 4.29$ \\
+ $\mathcal{L}_{curv}$  & $12.18 \pm 6.06$ & $9.98 \pm 6.22$ & $7.95 \pm 3.92$ \\
+ $\mathcal{L}_{cont}$  & $\mathbf{10.42 \pm 6.17}$ & $\mathbf{7.26 \pm 5.07}$ & $\mathbf{6.14 \pm 3.31}$ \\
\bottomrule
\end{tabular}
\end{table}

Regarding the components ablation study, in Tab.~\ref{tab:encoders_ablation}, we show \name~only with the in-view map encoder, adding the end points encoder, including the auxiliary topology head, and fine tuning the output. We observe how the additions help in better guiding the prediction, reducing RMSE errors in $16.97\%$, $38.55\%$ and $12.90\%$, respectively.

\begin{table}[b]
\centering
\caption{Ablation on Encoders \& Fine Tuning}
\label{tab:encoders_ablation}
\begin{tabular}{lccc}
\toprule
Method & RMSE (m) $\downarrow$ & CD (m) $\downarrow$ & FD (m) $\downarrow$\\
\midrule
In-view map & $17.62 \pm 10.59$ & $16.27 \pm 12.88$ & $13.63 \pm 8.42$ \\
+ End points & $14.63 \pm 9.77$ & $12.59 \pm 10.95$ & $10.65 \pm 6.87$ \\
+ Topology head  & $8.99 \pm 4.64$ & $11.65 \pm 7.21$ & $\mathbf{7.25 \pm 3.89}$ \\
\noalign{\vskip 3pt}
\hdashline
\noalign{\vskip 3pt}
+ Fine tuning & $\mathbf{7.83 \pm 3.66}$ & $\mathbf{7.08 \pm 4.94}$ & $7.48 \pm 2.95$ \\
\bottomrule
\vspace{-20pt}
\end{tabular}
\end{table}

\subsection{Real-time Evaluation}

Inference results are measured in a NVIDIA GeForce RTX 4060 GPU with batch size 1. Peak GPU memory consumption during inference is $221$MB  and average inference time is $15.3 \pm 1.2$ms, showing the real-time potential of the method.


\section{Conclusions}
\label{sec:conclusion}


\CG{We introduced the problem of beyond-view vectorized map generation for autonomous driving, being the first study to describe and explore the generation of vectorized map continuations beyond the observed field of view. Unlike existing works, which focus on reconstructing maps within the visible region, our formulation addresses the prediction of the map in unobserved areas. This problem aims to bridge the gap between short-sighted online methods and long-range offline ones by extending the vehicle horizon with plausible predictions. The experimental results show that this task is feasible under simplified road configurations, while also highlighting the challenges that remain for reliable map forecasting.}

\CG{As an initial approach to this problem, we presented \name, a Transformer-based implementation that provides a reproducible starting point for studying this task. Rather than targeting deployment-ready performance, \name~demonstrates the feasibility of the presented problem, identifies the main limitations, and provides a foundation for future research toward more complex driving scenarios.}

\subsection{Limitations and Future Work}

As a first study of beyond-view vectorized map generation, this work is intentionally limited to controlled scenarios in order to isolate the core challenges of the proposed task. The experimental evaluation focuses on simple road layouts and assumes accurate in-view maps, leaving the study of complex intersections, roundabouts and robustness for future work.


Although \name~is able to predict coherent road continuations in simple scenarios, the generated maps remain inaccurate at finer geometric and topological levels. In particular, the model struggles to correctly infer lane splits and merges, and often produces cluttered lane structures. These observations suggest that current architectures primarily capture local continuation patterns and that richer structural reasoning is required to model complex road topology.


Future work will investigate extensions that incorporate additional structural priors, such as SD map encoders similar to recent in-view map generation methods \cite{luo2024augmenting,jiang2024p}, together with improved loss formulations and lane-aware representations.  Extending the representation to richer HD map elements, including lane widths, road boundaries, and lane semantics, constitutes another promising research direction.






\section*{ACKNOWLEDGMENTS}
This work was primarily funded by the European Union’s Horizon Europe MSCA-2022-COFUND-01 under the Marie Skłodowska-Curie Grant 101126626. Additional support was provided by the MCIN/AEI/10.13039/501100011033 through the TRUEMAP project under Grant PID2024-162517OB-I00, also by MICIU/AEI/10.13039/501100011033 and the FSE+ under Grant JDC2024-055088-I.
	
	\bibliographystyle{IEEEtran}
	\bibliography{root} 
	
\end{document}